\documentclass[11pt]{article}

\usepackage[final]{acl}

\usepackage{times}
\usepackage{latexsym}
\usepackage[T1]{fontenc}
\usepackage[utf8]{inputenc}
\usepackage{microtype}
\usepackage{inconsolata}
\usepackage{graphicx}
\usepackage{booktabs}

\title{Fidelity Is Not Enough: Dispatch-Level Instrumentation\\
for Agentic Datasheet Extraction}

\author{
  Qing Ye \\
  Infineon Technologies AG \\
  Neubiberg, Germany \\
  \texttt{qing.leafnode@gmail.com} \\\And
  Meng-Hsuan Lin \\
  Infineon Technologies AG \\
  Neubiberg, Germany \\
  \texttt{hello.menghsuan@gmail.com}}

\begin{document}
\maketitle

\begin{abstract}
One model passed our fidelity check without ever opening the datasheet.
We found it while qualifying models for an internal extraction service:
a structured-output constraint had silently disabled tool use, and the
model answered anyway, with fabricated source text. Only the per-tool
trace exposed it. \emph{Fidelity}---whether an extracted value matches
the source---is the standard measure for agentic document extraction,
and it scores that run a success. We therefore log every tool call in
an agentic benchmark of 25 hand-curated claims over three components,
with 12 more on a fourth, 37 in all. From that \emph{dispatch record}
we build two instruments: a rule-based failure-attribution classifier,
and a silent-failure detector whose two rules check only which tools
were called, never the extracted value. The detector
raises no flag on 207 clean fidelity-passing extractions across three
model families, and recovers all 50 planted faults that withhold exactly
the tools its rules check. The two results are not symmetric: the first
bounds the false-positive rate, the second is recall by construction, and
detection power against runs that call their tools and still answer
wrongly is unmeasured. A second, independent oracle, a causal chamber
that tests whether the datasheet's claims hold under physical
measurement, is intentionally \emph{partial}: it confirms only what the
apparatus can exercise, a \emph{verifiable envelope} of 2 of those 37
claims, and we give a taxonomy of why the rest are not physically
gradable. Under a controlled perturbation, fidelity passes throughout
while the chamber verdict flips exactly at the measurement uncertainty.
Across three deployed model stacks (one destabilised by its serving
stack, not by any capability gap) the tool layer buys portability and
observability rather than accuracy, and earns its premium only once a
document outgrows the context window. We distil the lessons into a
checklist of cheap dispatch-level signals.
\end{abstract}

\section{Introduction}
\label{sec:intro}

Large language models are now routinely deployed as agents that extract
structured information from technical documents such as datasheets,
filings, and specifications. Most evaluations ask a single question:
does the extracted value match the document? That \emph{fidelity}
question matters, but because the score is computed against the
document the agent was given, it cannot distinguish a genuine
extraction from a correct-looking answer produced without reading it. It is also silent
on a second question deployment cares about: for claims checkable
outside the document, does the claim itself hold?

We hit exactly that failure while qualifying model and gateway
configurations for the datasheet parameter-extraction service behind
this work: an internal pilot, live since early 2026 behind a production
inference gateway, used on demand by engineers across multiple
divisions on our own and competitors' datasheets (126 extraction jobs to
date). A per-tool trace caught one model passing fidelity with
\emph{zero} tool calls, never opening the datasheet at all. A
fidelity-only benchmark scores this as success; only dispatch-level
instrumentation (a log of every tool call with its inputs, outputs, and
timing) makes it visible, and Section~\ref{sec:worked-example} traces
the cause. Instrumented experience is a strict subset of the pilot:
instrumentation was deployed part-way through and covers 17 of those
jobs, of which 14 are large-PDF and therefore eligible to raise a flag
at all. None has fired in live use; the zero-tool-call run itself was
found by a human reading a trace, not by a rule
(Appendix~\ref{sec:appendix-pilot}).

This paper presents a benchmark for agentic datasheet extraction
designed around that gap. The task is deliberately
narrow---\emph{targeted claim recovery}, not open-ended document
understanding---and each of 25 hand-curated quantitative claims, drawn
from the datasheets of three electronic components, is scored
on two separate outcomes: fidelity against the datasheet, and
reproducibility against an independent physical measurement from a
causal chamber. A further 12 claims on a fourth component test breadth
off corpus, 37 in all (Appendix~\ref{sec:appendix-fourth}). Our
contributions are:

\begin{itemize}
\item \textbf{Per-tool dispatch instrumentation} that turns the agent's
tool-call record into evidence for two instruments: a rule-based
failure-attribution classifier, and a silent-failure detector that flags
fidelity-passing runs that bypass navigation or skip verification, with
zero false positives on 207 clean extractions and by-construction recall
of all 50 planted faults
(Sections~\ref{sec:worked-example},~\ref{sec:res-attribution},~\ref{sec:res-silent}).
\item A \textbf{physical-grounding methodology}: each claim receives a
\emph{fidelity} verdict against the datasheet and a \emph{reproducibility}
verdict against an independent measurement from a causal chamber
\citep{gamella2025chambers}. Physically grounded evaluation is
necessarily \emph{partial}: an apparatus confirms only the claims its
instrumentation can exercise---its \emph{verifiable envelope}, 2 of those
37 claims---and our taxonomy of the unverifiable claims makes that
envelope explicit (Sections~\ref{sec:design},~\ref{sec:res-repro}).
\item A \textbf{cross-provider deployment study} across three model
families. It reports the cost, latency, and portability consequences of
running such an agent in production, distilled into a checklist of
cheap dispatch-level signals (Table~\ref{tab:lessons}). It also covers
documents that outgrow a context window, where agentic navigation stays
cheap because its cost tracks the pages a claim needs, not the
document's length (Section~\ref{sec:results},
Appendix~\ref{sec:appendix-stress}).
\end{itemize}

\section{Related Work}
\label{sec:related}

\paragraph{Agentic document understanding.}
A growing line of work casts document QA as an agentic, tool-driven
task. DocDancer \citep{zhang2026docdancer} trains an end-to-end document
agent; AgenticRAG \citep{suresh2026agenticrag} equips a reasoning model with
navigation tools close to ours, yet is scored only on document-internal
recall. DCI \citep{li2026dci} replaces retrieval with general-purpose
terminal tools, and \citet{sen2026grep} show that the harness itself
strongly shapes accuracy. In all of these the correctness
oracle is \emph{document-internal}
(Table~\ref{tab:related}, Appendix~\ref{sec:appendix-figures}).

\paragraph{Datasheet extraction.}
ModelGen \citep{wei2025modelgen} is the closest work in the
semiconductor-datasheet space, extracting compact-model parameter sets
with a multimodal agentic workflow. Two differences separate it from
ours. Its correctness oracle is a multimodal LLM that visually scores
fitting quality against device-characteristic plots (validation inside
an LLM), whereas ours is independent physical measurement outside any
model; and it scores a whole model at the device level, where we score
individual claims. AMSbench \citep{shi2025amsbench} scores multimodal
models on analog/mixed-signal understanding. Closest in task, DocEDA
\citep{chen2024doceda} pulls analog-circuit parameters from documents and
\citet{peng2025hemt} build ASM-HEMT SPICE models from datasheet
curves, but both validate against the datasheet itself, never against
independent measurement---the gap our oracle is designed to close.

\paragraph{External oracles and failure attribution.}
Chemistry agents such as Coscientist \citep{boiko2023coscientist} and
ChemToolAgent \citep{yu2025chemtoolagent} \emph{conduct} physical
experiments;
RWE-bench \citep{li2026rwebench} grounds clinical agents in a real
medical database. These validate against an external oracle, but none
verify \emph{document extraction} against \emph{physical measurement}.
The Causal Chambers \citep{gamella2025chambers} provide our
oracle (to our knowledge not previously used for document
understanding), and our document-side tool surface is our
released \texttt{datasheetindex} library \citep{datasheetindex}, after
the reasoning-over-structure design of PageIndex \citep{pageindex2025};
our contribution is the methodology, not the tools.
For process diagnosis, \citet{zhang2025whowhen}
pinpoint the decisive step in multi-agent failures, trajectory-aware
benchmarks score tool-selection over final answers
\citep{he2025trajectbench}, and surveys name fine-grained diagnosis as a
gap \citep{yehudai2025survey}, but these operate at per-agent or per-step
granularity; none, to our knowledge, attributes failures \emph{per tool}
against a fixed causal rubric.

\section{Benchmark Design}
\label{sec:design}

The benchmark mirrors the two questions in the introduction. It pairs a
document-internal \emph{fidelity} check on the agent's extraction with a
claim-level physical \emph{reproducibility} check from the chamber. Each
component inside a causal chamber \citep{gamella2025chambers} has a
manufacturer datasheet, and the chamber can independently measure a
subset of the quantities the datasheet claims, under matched conditions
where its instrumentation permits.

\subsection{Two verdicts}
\label{sec:design-verdicts}

We score two things, and keep them separate. \textbf{Fidelity} asks
whether the agent recovered what the datasheet \emph{says}: the value,
its units, and its stated operating conditions. \textbf{Reproducibility}
asks whether, with the chamber configured to match those operating
conditions, the measured value agrees with the stated value within a
spec-appropriate tolerance. The reproducibility verdict is computed by fixed
chamber-side code from the curated claim spec and the physical
measurement (neither of which comes from the agent), so it is identical
across runs and independent of the agent. The fidelity side's
independence is structural: the two-pass freeze
(Section~\ref{sec:design-agent}) withholds the chamber tools until the
extraction is submitted.

Reproducibility rests on one explicit assumption. We call an operating
condition \textbf{decisive} when the chamber must match it for the
reproducibility verdict to be meaningful; examples include supply
voltage, temperature, sampling rate, filter state, and
sample-population size. We report, per claim, which decisive
conditions could and could not be matched.

The chamber is an oracle, not a metrology lab: its measurements carry
instrument and model-form uncertainty. We attach an uncertainty estimate
to every measurement and treat any disagreement smaller than the
\textbf{combined} uncertainty $\sqrt{u^2+t^2}$ (cross-sensor
measurement uncertainty $u$ and the claim's own spec tolerance $t$) as
\emph{inconclusive} rather than a failure.

\subsection{The chamber and the claim set}
\label{sec:design-claims}

The 25 hand-curated quantitative claims span three components with
deliberately different chamber-side test shapes: 11 for the DPS310
barometer, 9 for the Si115x light sensor, and 5 for the ACS70331
current sensor. The DPS310 appears four times at distinct chamber
positions reading the same ambient pressure, which supports a
cross-sensor agreement protocol. The three Si115x devices sit at
non-equivalent positions, so its protocol uses per-channel precision
and range checks instead. The ACS70331, a discontinued part chosen
for its scarcity in pre-training corpora, serves mainly as a
fidelity-only check of whether the agent engages the document at all
(an \emph{engagement} diagnostic): its chamber-side verdict is returned
as inconclusive by design.

\paragraph{Claim selection.}
The gradable components are not a free choice: physical grounding
requires the chamber to measure the stated quantity, so that set is
exactly the apparatus's instrumented sensors. Within each component the
claims are hand-curated, not sampled: quantitative parameters with an
explicit value, unit, and operating condition---a diagnostic suite built
to exercise both verdicts, not a representative draw from each
datasheet's full parameter population
(Appendix~\ref{sec:appendix-claims}).

\subsection{The instrumented agent}
\label{sec:design-agent}

The agent runs in our own loop, not a third-party framework, because
the agent design is itself part of the experiment; the loop is a few
hundred lines, mostly instrumentation (trace logging, retry and timeout
control, token-and-call budgets).

It runs in two phases. The \emph{extraction} phase exposes only
document-side tools from \texttt{datasheetindex}
(\texttt{build\_datasheet} loads a datasheet and its contents;
\texttt{get\_section\_text}, \texttt{search\_text},
\texttt{extract\_table\_markdown}, and \texttt{inspect\_page} navigate
it), plus a \texttt{submit\_extraction} tool that \emph{freezes} the
value. Submitting it reveals the \emph{chamber} phase: five
chamber-side tools and a \texttt{submit\_chamber\_outcome} tool for
the agent's own, ungraded prediction of the measurement. We call this two-phase gating the
\emph{two-pass freeze}. Every call is recorded with its inputs, outputs,
and timing---the \emph{dispatch record} the rest of the paper builds on.

This per-tool instrumentation turns a pass/fail score into a diagnosis.
We commit up front to a fixed failure-attribution rubric
(Appendix~\ref{sec:appendix-rubric}): \emph{tool-output} errors are
charged to the returning tool, and \emph{tool-selection},
\emph{condition-omission}, and \emph{verification-skipped} errors to
reasoning. This keeps per-tool rates honest and exposes the share of
failures no tool fix would address.

\paragraph{The two silent-failure rules.}
The detector of Section~\ref{sec:res-silent} reads the same record
through two fixed rules, and both are predicates over tool
\emph{presence} in a single extraction that has already passed fidelity.
\emph{Tool-bypass} fires when that extraction made no navigation call.
\emph{Verification-skipped} fires when it made at least one navigation
call but never called the full-text search (\texttt{search\_text}) or
table-extraction (\texttt{extract\_table\_markdown}) tool. Those are the
tools an independent second look would use; the rule reads their
presence, not their role. Neither
rule reads the extracted value, the gold, or the chamber, and no
chamber-side call can satisfy either: both rules count datasheet-side
tools only.

\paragraph{Methodology audit.}
A late code review found two channels through which the agent prompt
leaked the graded answer. We closed both, pinned the fix with
regression tests, and re-ran the full matrix; every result in
Section~\ref{sec:results} is post-audit (Appendix~\ref{sec:appendix-audit}).

\section{Experimental Setup}
\label{sec:setup}

We evaluate three models from independently trained families: Claude
Sonnet~4.6, GPT-5.1, and Qwen3.6-27B, an open-weights model standing in
for a third provider (its headline fidelity run is
pre-freeze).\footnote{A backend change
(Appendix~\ref{sec:appendix-portability}) blocked re-running Qwen under
the two-pass freeze; re-running the frontier models under it left their
fidelity unchanged (Section~\ref{sec:res-fidelity}). Qwen's
reproducibility verdict, computed chamber-side, is unaffected either
way.} All
three run the \emph{same} agent harness (loop logic, tool surface,
prompts, and budgets of Section~\ref{sec:design}), but not the same
transport: Claude and Qwen run through an Anthropic-shaped gateway
endpoint, GPT-5.1 through the OpenAI Responses API. A uniform wire format
silently distorts the models it was not designed for, so we hold the
harness \emph{logic} fixed, not the transport. Cross-model differences
therefore reflect a model-stack configuration, not weights alone
(Section~\ref{sec:res-portability}).

Each of the 25 claims is run under two engines. The \emph{agentic}
engine uses the full tool surface of Section~\ref{sec:design} under a
uniform 30-turn budget fixed in advance. The \emph{single-pass baseline}
sends the entire datasheet PDF in one call. The baseline runs end-to-end
on all three models; because the vLLM backend hosting Qwen cannot ingest
a PDF, Qwen's baseline renders every page to an image
(Section~\ref{sec:res-portability}).

To separate run-to-run variance from genuine cross-model differences, we
run the agentic engine three times per claim per model. Each (model,
claim, repeat) extraction is a \emph{cell}, the unit of the evaluation
matrix (Appendix~\ref{sec:appendix-cells}). Qwen's repeats ran
2026-05-21/22, before the two-pass freeze, and the Claude and GPT-5.1
repeats 2026-06-05 under it, all after the audit and
the transport fix
(Sections~\ref{sec:design},~\ref{sec:res-portability}); we report a
mean\,$\pm$\,std over them. Sampling is left at each provider's defaults
(no fixed temperature, top-$p$, or seed); only the loop logic, reasoning
effort (Claude and GPT-5.1 at \texttt{medium}; Qwen's stack offers only
on/off, left on), and turn budget are pinned. The repeats therefore
measure the end-to-end variance of a deployed configuration, not
sampling noise.

\section{A Worked Example: A Silent Failure Caught in Development}
\label{sec:worked-example}

The zero-tool-call failure of Section~\ref{sec:intro} is worth tracing
in full, because it generalises to any agent given a structured-output
constraint. Qwen passed fidelity with \emph{zero} tool calls on every
claim. To guarantee a well-formed answer, the loop set each request's
\texttt{output\_format} parameter---not a tool---to a JSON schema. On
the Anthropic models the schema binds only the terminal turn, so tool
use is unaffected; on the open-weights backend the same field became a
guided-decoding token mask. The mask fixed the output to the answer
schema from the first token, blocking the very token that begins a tool
call. Forced into the answer format, the model filled it without
reading, emitting a fluent value it never extracted from the document,
complete with fabricated source text.

Fidelity scores this a success: the value was right. Only the dispatch
record exposes the failure, and it points to the fix. We removed that
parameter and delivered structured output through a tool call instead,
which composes uniformly across providers:
\texttt{submit\_claim\_result}, the single-pass predecessor of
Section~\ref{sec:design-agent}'s \texttt{submit\_extraction}.
Figure~\ref{fig:engagement} (Appendix~\ref{sec:appendix-figures}) tracks
the effect across three development revisions (on the then-deployed
Qwen3.5-27B); under the fix, Qwen engages the datasheet and exercises
the tool surface comparably to the others (Figure~\ref{fig:dispatch}).
Every result in Section~\ref{sec:results} is measured under the fix.

The point is not the misconfiguration itself but that the benchmark
makes silent, harness-level failures observable: any answer a model
never extracted leaves the same fingerprint, missing navigation or
cross-check calls (Section~\ref{sec:res-silent}).

\section{Results}
\label{sec:results}

With the \texttt{output\_format} bypass of
Section~\ref{sec:worked-example} fixed, we report the post-audit,
provider-native runs. Fidelity is high for the frontier models;
run-to-run stability is what separates the three
(Appendix~\ref{sec:appendix-rescore} re-scores fidelity under
exact-value matching).

\subsection{Fidelity and run-to-run variance}
\label{sec:res-fidelity}
\begin{table}[t]
\centering
\small
\setlength{\tabcolsep}{4pt}
\begin{tabular}{@{}lccc@{}}
\toprule
\textbf{Model} & \textbf{Fidelity} & \textbf{Eng.\ err.} & \textbf{Latency} \\
\midrule
Claude Sonnet 4.6 & 25.0\,$\pm$\,0.0 & 0/0/0 & 76\,$\pm$\,5\,s \\
GPT-5.1           & 25.0\,$\pm$\,0.0 & 0/0/0 & 236\,$\pm$\,78\,s \\
Qwen3.6-27B       & 19.0\,$\pm$\,4.0 & 1/6/10 & 130\,$\pm$\,2\,s \\
\bottomrule
\end{tabular}
\caption{Agentic-engine results across three deployable model-stack
configurations, over three repeated runs. Fidelity is the pass count of
25; the engine-error column gives the per-repeat counts (a timeout, or a
turn ending without the submit call). Qwen's instability is
reasoning-mode-dependent: 25/24/24 with reasoning disabled
(Appendix~\ref{sec:appendix-qwen-thinking}). Latency is a wall-clock mean over
cells that returned a result, dominated by time outside local tool
execution (Section~\ref{sec:res-cost}); $\pm$ values are standard deviations
across the three repeats.}
\label{tab:results}
\end{table}

Under the agentic engine the two frontier models are both accurate and
run-to-run stable: Claude Sonnet~4.6 and GPT-5.1 each pass all 25 claims
in every repeat with no engine errors (Table~\ref{tab:results}). This
holds under the two-pass freeze (Section~\ref{sec:design-agent}), which
makes fidelity independence structural rather than a prompt instruction.
A trace audit shows what the freeze changed: before it, 80\% of Claude
cells and 100\% of GPT-5.1 cells called a chamber tool before
submitting their extraction (92\% in Qwen's reported run, which predates
the freeze); under it, none can. Qwen3.6-27B is the
unstable model: it passes 23, 19, and 15 claims across the three repeats
(mean $19\pm4$). A different subset of cells ends without a submit call
in each run, so only 13 of 25 claims keep the same verdict across
repeats, against 25 of 25 for the frontier models. This is a documented
vLLM/reasoning-mode interaction \citep{qwen3issue1817}, a
deployment-stack artefact rather than a capability gap: when re-run with
reasoning disabled (the workaround the upstream issue recommends), Qwen
passes 25/24/24 of the same claims with a single engine error
(Appendix~\ref{sec:appendix-qwen-thinking}). Per-claim fidelity is in
Figure~\ref{fig:fidelity}, confidence (near-saturated, not validated) in
Figure~\ref{fig:confdist}.

\paragraph{A fourth component, off corpus.}
Breadth beyond the three: 12 claims on a 20-page A4988
stepper-driver datasheet are scored on fidelity only. Both frontier
models pass 12 of 12, under exact-value matching too, with no engine
errors and no detector flags, and a re-run on a changed tool surface
still passes every claim it completes. We report it as a null result on
an already-exposed document, not generalisation. That document is
navigated measurably more cheaply, its claims have no declared candidate
pool, and its PDF was the decoy in the probe arm of
Section~\ref{sec:res-silent}, so both models had been seen
transcribing some of its rows before the claims were written. Zero
flags over 12 claims is a zero-event count, not a precision result
(Appendix~\ref{sec:appendix-fourth}).

\subsection{Cross-provider portability and industry lessons}
\label{sec:res-portability}
The same agent, run through one inference gateway against three model
families, surfaced six portability failures, two of them \emph{silent}:
the tool-bypass, and a GPT-5.1 configuration that dropped its reasoning
trace without error.
Reaching GPT-5.1 through its provider-native Responses API resolves the
two GPT-5.1 failures (one loud, one silent); the other four are backend
limitations (Appendix~\ref{sec:appendix-portability}).
Table~\ref{tab:lessons} distils the pattern into an operational
checklist: each cheap dispatch-level signal, the failure it exposes
(most return HTTP success), and the fix it enabled. The lesson:
provider portability is not settled by a working SDK call; it often
fails downstream, often silently, and only instrumentation separates
that from a real pass.

\begin{table}[t]
\centering
\footnotesize
\setlength{\tabcolsep}{2pt}
\begin{tabular}{@{}lll@{}}
\toprule
\textbf{Signal logged} & \textbf{Failure it exposes} & \textbf{Action enabled} \\
\midrule
Dispatch record   & fidelity pass, 0 calls & output via tool call \\
Reasoning trace   & reasoning dropped           & provider-native API \\
HTTP vs.\ call    & 200 OK, no call             & alert on backend \\
Rubric label      & tool vs.\ reasoning         & fix tool or prompt \\
Cross-check call  & verification skipped        & flag for human review \\
\bottomrule
\end{tabular}
\caption{Industry lessons. The first two rows are the \emph{silent}
portability failures. All five are warning signals that route a run to
human review, not automatic release gates; Section~\ref{sec:res-silent}
reports what the cross-check signal does and does not certify.}
\label{tab:lessons}
\end{table}

\subsection{Baseline versus agentic}
\label{sec:res-cost}
On this corpus baseline and agentic agree on fidelity in 68 of 75 cells
(25 claims $\times$ 3 models).
The reproducibility verdict, computed chamber-side, is engine-invariant
by construction: baseline and agentic differ on one Qwen cell only,
and only because that cell was lost to an engine error. The tool layer
charges a small premium over the single-pass baseline at list prices, $1.2\times$ for Claude and
$1.8\times$ for GPT-5.1; Qwen's $3.9\times$ reflects a cheap page-image
baseline rather than an expensive agent (Figure~\ref{fig:cost} gives the
absolute figures). Latency is dominated by time outside local tool
execution (generation plus gateway queueing), not a ranking
(Table~\ref{tab:results}, Appendix~\ref{sec:appendix-latency}).

\paragraph{The operational rule.}
Default to single-pass; switch to agentic only when a document exceeds
the context window or a provider's page cap. Here every datasheet fits
one window (at most 65 pages), so single-pass is correct: the tool layer
buys portability and observability, not accuracy. That flips on a
397-page PMIC datasheet, where agentic navigation stays bounded-cost and
single-pass ingestion does not, reading just the pages each claim needs
(Appendix~\ref{sec:appendix-stress}).

\subsection{Failure attribution}
\label{sec:res-attribution}
The per-tool dispatch record feeds two instruments. The first diagnoses
runs that \emph{failed}: a failure-attribution classifier---a
deterministic rule set, no model in the loop---labelling every tool call
against the fixed rubric of Section~\ref{sec:design-agent}. Neither
frontier model has a single flagged call; Qwen3.6-27B has 41, but most
of those are steps the rubric could not label rather than real
failures, because Qwen's reasoning-enabled traces expose more steps. We validate it on 30 traces stratified across models and
hand-labelled blind to its predictions
(Appendix~\ref{sec:appendix-rubric}). Three drew an abstention, leaving
27 adjudicated cells: 21 the human called clean, 6 problematic. It
agrees on 24 of the 27 (89\%, Cohen's $\kappa=0.61$); all three
disagreements are cases where it was too \emph{lenient}, and no
over-flag was observed. The sample is small: zero over-flags on 21 clean
cells is still consistent with a true over-flag rate of up to about
13\% (one-sided 95\% bound), and catching 3 of 6 problematic cells
says little about sensitivity. So the classifier is directional only;
it cannot attribute Qwen's higher flag count to genuine process
differences, a reading we leave open
(\hyperref[sec:limitations]{Limitations}).

\subsection{Detecting silent failures}
\label{sec:res-silent}
The second instrument targets the worked example's failure: a run that
passes fidelity but reaches its answer through a defective
process---what \citet{cao2026corruptsuccess} call \emph{corrupt
success}. Two fixed rules read a fidelity-passing cell's dispatch
record: \emph{tool-bypass} (no navigation call) and
\emph{verification-skipped} (navigation, no cross-check); neither reads
the extracted value. We plant silent failures by fault injection. For
each of the 25 claims we make two Claude Sonnet~4.6 runs, one with
navigation tools withheld and one with cross-check tools withheld, and
pin the value to a fidelity-passing reference; each of the 50 runs is
therefore a silent failure by construction. Fidelity-only scoring flags
none of the 50; the detector recovers all 50. What makes it a usable
warning, not a tautology: it also flags none of the 207 fidelity-passing
cells of ordinary runs (Table~\ref{tab:silent}).
The 207 cells are not independent trials: they come from only 74
claim-by-model groups ($25\times3$, less one Qwen claim that never
passed) and 25 claims. Zero flags then bounds the false-positive rate
(one-sided 95\%) at 4.0\% per group or about 11\% per claim, not the
1.4\% that treating all 207 cells as independent would give.
Under the same injection, the detector recovers every
silent failure on GPT-5.1 and Qwen3.6-27B (32/32 and 21/21), so recall
holds across all three; the smaller denominators count only completed
silent failures: many injected GPT/Qwen runs abort loudly, and two Qwen
runs had no fidelity-passing reference to pin. Because the planted
faults withhold exactly the tools the rules check, recall is by
construction and limited to these rule-aligned faults, not silent
failures in general (\hyperref[sec:limitations]{Limitations}).

\paragraph{What the rules miss.}
Both rules read tool \emph{presence}, so a run that uses its tools on
the wrong document defeats them. We measure that case: two probe arms leave navigation, cross-check and read
success intact and corrupt only the grounding, one returning empty
content from every call and one serving every call from a decoy
datasheet. Across their 100 runs on Claude and GPT-5.1, 8 answered and
none was right. (Qwen3.6-27B is excluded: it errors on 35 of its own 50
probe runs under forced tool choice;
Appendix~\ref{sec:appendix-probes}.) Six of the 8 are faithful readings
of the decoy---Claude transcribed a motor-driver table row verbatim
against an Si115x operating-temperature claim---and only 2 are memory
answers. On the decoy arm, a content check (is the quoted
source text findable in the intended datasheet?) catches the class but
cannot tell its runs apart. The arm stored a quotation in 11 runs (6
that answered, 5 that declined); none is findable in the intended
datasheet, so the check flags all 11, but only 2 are verbatim spans of
the decoy, so it separates 2 of the 6 answering runs; the other 4
paraphrase the decoy, which exact matching would miss whether the
answer were right or wrong. Only 2 memory answers were observed, too
few to bound how often that case occurs or is caught (a denominator of
2 puts the one-sided 95\% ceiling at 77.6\%); it stays open.

\paragraph{Benchmark and deployment differ.}
The benchmark and production detectors are not the same instrument,
and results do not transfer between them. The benchmark evaluates verification-skipped, a rule the
production service does not ship; production ships two the benchmark
does not evaluate: a tool-read-failure rule, and a source-grounding pass
that raises a flag only when the locator call itself errors, so a
genuine failure to re-find a quoted value raises nothing. A clean flag
sheet from the pilot therefore carries no information about the
grounding check.

\begin{table}[t]
\centering
\small
\begin{tabular}{@{}lrrr@{}}
\toprule
\textbf{Planted fault} & \textbf{N} & \textbf{Fidelity-only} & \textbf{Detector} \\
\midrule
Tool-bypass          & 25 & 0 & 25 \\
Verification-skipped & 25 & 0 & 25 \\
\bottomrule
\end{tabular}
\caption{Planted silent failures recovered (Claude Sonnet~4.6).
Fidelity-only scoring catches none; the detector catches all 50 at zero
false positives on 207 fidelity-passing cells spanning all three model
families (the post-audit reference run plus two further repeats). Recall
on
these rule-aligned injections is by construction
(Section~\ref{sec:res-silent}).}
\label{tab:silent}
\end{table}

\subsection{The reproducibility decomposition}
\label{sec:res-repro}
Fidelity is only one of the two verdicts. With current chamber
instrumentation only 2 of the 25 frozen claims reach a definitive
\emph{pass} and none a definitive \emph{fail}; the other 23 are
\emph{inconclusive}. Inconclusive is a decision rule, not indecision.
The fourth component adds no gradable claim at all, so the envelope over
the combined corpus is 2 of 37; the taxonomy below is stated over the
frozen 25, which are the only claims a chamber protocol was ever
designed for. The two passes are weak range-membership checks: measured
pressure lies inside the DPS310's stated operating range, and Si115x
ADC codes lie inside the range its stated bit depth implies. The substantive finding is therefore the \emph{negative result}
itself: a taxonomy of why the other 23 claims are not physically
gradable with the available instrumentation. Five are
\emph{engagement-only} (the ACS70331 leg the chamber never stages). Five
are \emph{absolute-accuracy}, ungradable without a traceable reference.
Another 12 depend on a decisive condition the chamber cannot match (a
VDD sweep, a held temperature, a device-internal quantity). One is
\emph{resolution-limited} ($\approx$0.080\,hPa cross-sensor uncertainty
vs.\ a $\pm$0.06\,hPa tolerance). None is an apparatus failure: the
verdicts are deterministic chamber-side, and
Appendix~\ref{sec:appendix-inconclusive} works one claim from each
class.

\begin{figure}[t]
\centering
\includegraphics[width=\columnwidth]{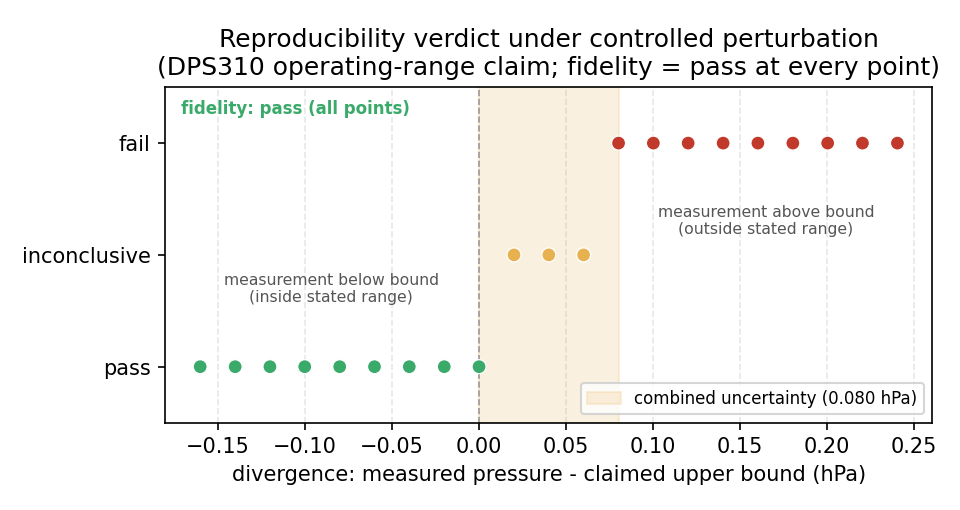}
\caption{Reproducibility verdict under controlled perturbation of the
DPS310 operating-range claim: the real cross-sensor measurement
($945.285\pm0.080$\,hPa) is held fixed while the datasheet-stated bound
is shrunk through it; the verdict flips exactly at the combined
uncertainty (shaded). The perturbation is a controlled vehicle, not a
claim that the datasheet is in error.}
\label{fig:perturb}
\end{figure}

\paragraph{The oracle under controlled divergence.}
The corpus itself contains no definitive reproducibility failure, so we
manufacture a calibrated one: holding the real DPS310 pressure fixed,
we shrink the datasheet's stated bound through it. The perturbed
datasheet runs end-to-end on Claude Sonnet~4.6: fidelity passes
throughout, while the reproducibility verdict walks
pass\,$\rightarrow$\,inconclusive\,$\rightarrow$\,\emph{fail} exactly at
the $0.080$\,hPa uncertainty band (Figure~\ref{fig:perturb}).

\section{Conclusion}
\label{sec:conclusion}
Fidelity is necessary but not sufficient for trusting an agentic
extraction system. What we would carry to another deployment is the
dispatch record, which exposes silent failures that fidelity-only
scoring passes. The physical oracle contributes something narrower: a
verifiable envelope stated before any reproducibility result, and a
taxonomy of why most claims fall outside it. Operationally: instrument
every tool call, and use the cheap dispatch-level signals to route a
suspect run to human review rather than to gate a release automatically
(Table~\ref{tab:lessons}). Go agentic only once a document outgrows the
context window or page cap.

\section*{Limitations}
\phantomsection\label{sec:limitations}
\paragraph{False positives, not detection power.}
The zero-false-positive result bounds how often the rules fire on clean
runs: on 207 clean cells, never. It does not measure detection power. The
cross-check that verification-skipped requires is present on 207 of 207
clean cells, so ``required by everything'' and ``flagged nothing'' are
one fact here, not opposite ones, and the rule fires only on a
catastrophic condition. Every fault arm producing positives also
withholds the tools those rules read, so measured recall stays
structural.

\paragraph{Detection power on degraded runs is unmeasured.}
Both rules are silent by construction on an agent that calls its tools,
reads them successfully, satisfies the predicate and still answers from
the wrong content, and in production fidelity is silent on it too.
Section~\ref{sec:res-silent} shows that class is real rather than
hypothesising it; nothing here separates it from a clean run.

\paragraph{No human baseline.}
We do not compare the agent against an applications engineer reading the
same datasheet, so how hard these claims are for a competent human
remains open; the omission is deliberate. The blind re-derivation of Appendix~\ref{sec:appendix-rederivation} bounds the required substrings, not the difficulty.

\paragraph{Small-sample failure attribution.}
The classifier is validated against a single blind annotator on
$n{=}27$ adjudicated cells ($\kappa=0.61$), and the per-class
denominators are what bound it: no over-flag observed on 21 human-clean
cells is a one-sided 95\% ceiling near 13\%, not a demonstration that it
never over-flags, and sensitivity is 3 of 6
(Appendix~\ref{sec:appendix-rubric}). Nothing here shows the classifier
is calibrated.

\paragraph{Reproducible, given a gateway.}
The benchmark, claim file, grading surface, archived runs and harness are
released \citep{datasheetindex}. We cannot ship a model: each arm runs
through a gateway the replicator supplies, so numbers reproduce only up to
checkpoint drift.

\paragraph{Design provenance.}
Every Section~\ref{sec:results} result was re-run after the oracle leak
was found, but most of the
\emph{design} predates the fix: the claim set, its required substrings,
its confidence floors and the attribution rubric were all fixed while
the prompt builder was leaking the answer and the grading tolerance, and
a threshold chosen because leaky runs passed it stays contaminated after
a re-run. The detector rules are the exception, first authored four days
\emph{after} the fix (Appendix~\ref{sec:appendix-audit} dates each
artefact). A blind re-derivation now bounds the required substrings; its
confidence-floor half was lost to ambiguous instructions, so only the
archive bounds those (Appendix~\ref{sec:appendix-rederivation}). The
claim set and its conditions stay unbounded.

\paragraph{The chamber-side prediction is ungraded.}
The second-phase \texttt{submit\_chamber\_outcome} field is a sanity
signal, not a scored quantity; no result depends on it. The two-pass
freeze (Section~\ref{sec:design-agent}) closes the concern that nothing
structurally prevented chamber data from reaching the datasheet-side
value.

\paragraph{The verifiable envelope is an apparatus limit.}
Only 2 of the 37 claims are physically gradable, and none of the fourth
component's 12 can be staged at all. The inconclusive taxonomy
(Appendix~\ref{sec:appendix-inconclusive}) doubles as a requirements
list for instrumentation a production test laboratory already operates.

\paragraph{Cross-model confounds.}
The three configurations differ in transport, input modality, freeze
state and reasoning mode as well as in weights, so this compares
deployed stacks, not models; Qwen's instability in particular is largely
a deployment-stack artefact (Appendix~\ref{sec:appendix-qwen-thinking}).

\paragraph{Scope.}
The corpus is 25 claims across three components on one inference
gateway, plus 12 off corpus. The portability findings are properties of that
operational surface, not in-principle claims about any provider, and the
cost figures use public list prices.

\section*{Ethical Considerations}
The benchmark is built entirely from public assets: the Causal Chambers'
open hardware, data, and simulators, and publicly available manufacturer
datasheets. No proprietary information is used or released. A
fidelity-passing extraction is explicitly \emph{not} a guarantee that a
datasheet claim is physically true; the reproducibility axis exists so
that this distinction is not lost. Practitioners deploying
document-extraction agents should treat extracted values as faithful
transcriptions to be verified, not as validated facts.

\bibliography{custom}

\appendix

\section{Full Cross-Provider Portability Findings}
\label{sec:appendix-portability}

Table~\ref{tab:portability} summarises six portability failures observed
on the internal LiteLLM gateway; we expand each here.

\begin{table}[t]
\centering
\small
\resizebox{\columnwidth}{!}{%
\begin{tabular}{@{}lll@{}}
\toprule
\textbf{Engine / backend} & \textbf{Failure mode} & \textbf{Detect.} \\
\midrule
Baseline, GPT-5.1 passthrough & PDF block rejected   & Loud \\
Agentic, GPT-5.1 passthrough  & reasoning dropped    & \textbf{Silent} \\
Baseline, Qwen (vLLM)         & PDF not ingestible   & Loud \\
Agentic, Qwen (vLLM)          & tool tokens masked   & \textbf{Silent} \\
Agentic, DeepSeek (vLLM)      & tool tokens absent   & Loud \\
Agentic, Llama (vLLM)         & replies unparseable  & Loud \\
\bottomrule
\end{tabular}%
}
\caption{Six cross-provider portability failures surfaced on one
inference gateway. Two are \emph{silent} (guided-decoding tool masking
and dropped reasoning) and are the ones fidelity-only scoring would
miss. The two GPT-5.1 rows are artefacts of the Anthropic-shaped
passthrough and are resolved by the provider-native Responses API
(Section~\ref{sec:setup}); the rest are backend limitations.}
\label{tab:portability}
\end{table}

\paragraph{(1) Single-pass baseline, GPT-5.1 passthrough.}
The baseline engine sends the datasheet as a native PDF document content
block. Routed through the gateway's Anthropic-shaped endpoint to the
Azure-hosted GPT-5.1, the block is translated into a malformed image
payload, which Azure rejects with an HTTP~400 on every cell. The
provider-native Responses API accepts a PDF directly, and the baseline
then runs end-to-end.

\paragraph{(2) Agentic, GPT-5.1 passthrough: silent reasoning loss.}
The same Anthropic-shaped passthrough down-translates an agentic request
to a chat-completions call. The agent runs to completion and passes
fidelity, but the provider's reasoning output is discarded in
translation: the recorded reasoning trace is empty, with no error
raised. The provider-native Responses API preserves
reasoning summaries, and all GPT-5.1 numbers in this paper are measured
under it.

\paragraph{(3) Single-pass baseline, Qwen (vLLM).}
The vLLM backend hosting Qwen cannot ingest a PDF at all; an
Anthropic-native \texttt{document} block is rejected upstream. Unlike
(1), this is a backend limitation, not a passthrough artefact, so we
render every datasheet page to an image and send those; the VL-capable
Qwen reads them, and the baseline runs end-to-end.

\paragraph{(4) Agentic, Qwen: the silent tool-bypass.}
As described in Section~\ref{sec:worked-example}, the JSON-schema
\texttt{output\_format} constraint is realised by vLLM as guided
decoding, whose token mask suppresses the tool-call opening token. The
agent emits a schema-valid answer with zero tool calls. This is the only
one of the six that produces a fidelity pass on an answer the agent
never extracted.

\paragraph{(5) Agentic, DeepSeek-R1-70B.}
The served tokenizer does not expose the tool-call delimiter tokens the
vLLM tool parser expects; every request is rejected upstream.

\paragraph{(6) Agentic, Llama-3.3-70B.}
The model returns responses that the loop's Anthropic-shaped parser
classifies as neither a tool call nor a terminal answer; the loop
exhausts its turn budget with no progress.

These findings describe one concrete operational surface at one point
in time, not any provider in principle.

\paragraph{Portability is not static: a mid-study backend change.}
After the runs above, a reliability-oriented change to the hosted Qwen3.6-27B backend regressed
its agentic tool-calling: with a large tool schema (our $3.7$\,kB
dereferenced extraction-result schema) the model stops emitting a
structured tool call and instead repeats a tool-call token until the
output limit, so the loop sees a terminal turn with no call. The same
model handles small schemas cleanly, and forcing constrained decoding
recovers it only partially, so the fault is the backend's handling of
large schemas, not the model or our API shape: it reproduces on both
the Anthropic-shaped passthrough and the native chat-completions
endpoint. Each call still
returns HTTP~200, so the regression is invisible without the per-call
dispatch record: the same lesson as~(2) and~(4), now across \emph{time}
rather than across providers. We therefore report Qwen's fidelity from
the pre-change run (Section~\ref{sec:res-fidelity}). The two-pass
freeze of Section~\ref{sec:setup} is itself verified on Claude and GPT-5.1 and on a sample of Qwen two-pass
traces, all with $0\%$ pre-submit chamber-tool calls.

\section{Failure-Attribution Rubric}
\label{sec:appendix-rubric}

Per-tool error rates are meaningful only if each failure can be assigned
to a specific cause. The benchmark commits in advance to a fixed rubric:

\begin{itemize}
\item \textbf{Tool-output error}: a tool returned wrong content (for
example, a mangled table row). Attributed to that tool.
\item \textbf{Tool-selection error}: the right tool existed but the
agent called the wrong one, or none. Attributed to reasoning.
\item \textbf{Condition-omission error}: the agent extracted a value but
dropped or misread its operating conditions. Attributed to reasoning;
the missed condition is logged.
\item \textbf{Verification-skipped error}: the agent finalised without
exercising its cross-check step. Attributed to reasoning.
\end{itemize}

Only tool-output errors are charged to a tool; the other three are
reasoning errors between tool calls. \emph{Verification-skipped} names
both a class here and a detector rule in
Section~\ref{sec:res-silent}, and the two are deliberately not the same
predicate: this class labels runs that \emph{failed} fidelity and counts
a chamber cross-sensor check as verification, while the detector rule
reads only fidelity-\emph{passing} runs and only datasheet-side tools.

\paragraph{How much room the cross-check predicate leaves.}
The detector rule is satisfied by a single qualifying call, so it is
worth reporting how far from that floor a clean run sits. Across the 207
clean cells the two cross-check tools are called 5.0 times per cell on
average (\texttt{search\_text} alone, 4.08), and every cell calls at
least one. What bounds the rule is the low tail, not the mean: only 6
of the 207 clear the predicate by exactly one call, 26 by two or fewer.
This distribution is the evidence for the Limitations claim that a
predicate every run satisfies with room to spare can fire only on a
catastrophic condition.

\paragraph{What the classifier is, and what ``41'' counts.}
The classifier is a deterministic rule set, not a model. It walks the
dispatch record and assigns every tool call---and the final output, when
the cell failed---to one of the four rubric slots above, to \texttt{ok}
when the step sits in a successful chain, or to \texttt{unclassified}
when no rule applies. An optional LLM pass over the residual
\texttt{unclassified} steps exists but is \emph{off} for every number
in this paper, the validation below included, so the annotator scored
the rule set alone.

The 41 non-\texttt{ok} slots that Section~\ref{sec:res-attribution}
reports for Qwen3.6-27B are counted per \emph{call}, not per cell, and
they are not 41 rubric failures, and they fall in 7 of that model's 25
agentic cells. Of the 41, 21 are \texttt{unclassified}, the residual
slot where no rule applied, and 14 are \texttt{engine\_error}, a
harness outcome rather than an attribution. Six are rubric labels
proper---five verification-skipped and one condition-omission---in 6
cells.
Claude Sonnet~4.6 and GPT-5.1 carry no non-\texttt{ok} slot on any
agentic cell. The gap between 41 and 0 is thus dominated by steps the
rubric could not label, which is why we read it as directional rather
than as a failure count.

To check the classifier that applies this rubric, a 30-trace sample
stratified across models (10 per model) is hand-labelled by an
annotator independent of the classifier's development and blind to its
predictions. The annotator judges each trace at the whole-cell level,
recording the rubric label the classifier should have produced or
abstaining when the record alone is insufficient. The classifier's
per-call output is reduced to the same level by a fixed rule: a cell is
problematic if any call carries a non-\texttt{ok} label or the run hit
an engine error, and its label is that call's; otherwise it is clean.
Of the 30 traces the annotator abstained on three (all on the same
\texttt{si115x-adc-bit-depth} claim, where the bit depth is contextually
inferable but never surfaced in the extracted value), which we exclude.
On the remaining 27 adjudicated cells the classifier matches the human
label on 24 (89\%). The three disagreements are all classifier
\emph{misses} rather than over-flags (two tool-selection, one
verification-skipped); two fall on Claude Sonnet~4.6 traces and one on
Qwen3.6-27B. Because most cells are clean, raw agreement overstates
reliability, so we also collapse each rater's
call to a binary clean-versus-flagged decision, which gives 21 cells
both raters call clean, 3 both call flagged, 3 the classifier alone
calls clean, and none the classifier alone calls flagged. That yields
Cohen's $\kappa=0.61$ (observed agreement $88.9\%$, chance $71.6\%$).
Read the per-class denominators, not the headline: no over-flag was
\emph{observed}, but only 21 cells are human-clean, so the one-sided
95\% ceiling on the over-flag rate is near 13\%; sensitivity on the
6 human-problematic cells is 3 of 6. Six positives cannot support a
per-failure-type confusion table, so we report the aggregate only.

\paragraph{Annotation provenance.}
The annotator is the paper's second author, and the sequence matters.
The annotation was completed before they had any involvement with this
work: they had not read the paper, were blind to the classifier's
predictions, and took no part in its development. Co-authorship followed
the annotation and for unrelated contributions. A single annotator
remains a real limit whether or not they are independent. Both figures
are recomputed from the stored adjudication rather than transcribed.

\section{Oracle-Leak Audit}
\label{sec:appendix-audit}

A code review late in development found two channels through which the
agent prompt leaked the graded answer.

\paragraph{Channel 1: claim-spec serialisation.}
The prompt builder serialised the entire claim specification into the
agent's message, including the source page, the verbatim source text,
the substrings the answer had to contain, the claimed value, and the
scoring tolerance. The agent was, in effect, told the answer and how
leniently it would be graded.

\paragraph{Channel 2: curated descriptions.}
Curators had paraphrased answers into a free-text description field;
most claims leaked the numeric answer this way, and several leaked the
source page.

\paragraph{Fix.}
We replaced the serialisation with an allowlist projection that exposes
only an explicitly enumerated set of disambiguating fields, routed both
prompt builders through a single chokepoint so they cannot diverge,
rewrote the leaking descriptions, and restated one claim whose stated
operating condition was numerically equal to its claimed value; 12
regression tests pin the contract. The full matrix was re-run under the
fixed prompt on 2026-05-18, and again on 2026-05-20 after the transport
correction of Section~\ref{sec:res-portability}; the Claude and GPT-5.1
agentic numbers in this paper are from a further 2026-06-05 re-run under
the two-pass freeze, the GPT-5.1 single-pass baseline from that
2026-05-20 run, and the Qwen3.6-27B numbers from the 2026-05-21 re-run
(Section~\ref{sec:setup}), all post-audit and provider-native.

\paragraph{Design provenance, dated.}
Re-running the numbers does not re-make the decisions. Table~\ref{tab:provenance}
dates each artefact the results depend on against the fix, from the
repository history rather than from recollection.

\begin{table}[ht]
\centering
\small
\begin{tabular}{@{}llc@{}}
\toprule
\textbf{Artefact} & \textbf{Fixed} & \textbf{Rel.\ fix} \\
\midrule
Claim set, bounds, conditions   & 05-05\,--\,05-12 & before \\
Required-substring needles      & 05-05\,--\,05-12 & before \\
Confidence floors               & 05-05\,--\,05-12 & before \\
Attribution rubric, classifier  & 05-06            & before \\
\midrule
Prompt-leak fix                 & \textbf{05-18}   & --- \\
\midrule
Silent-failure detector rules   & 05-22            & after \\
Cross-check predicate narrowed  & 06-05            & after \\
Two-pass freeze                 & 06-05            & after \\
\bottomrule
\end{tabular}
\caption{When each artefact was fixed, relative to the prompt-leak fix
of 2026-05-18 (all dates 2026). The three grading-surface rows are the
exposure: they predate the fix and were never revisited. The detector
rules do not share it, having been authored four days afterwards, and
the 2026-06-05 revision made the predicate \emph{stricter} by removing
the chamber-side cross-check.}
\label{tab:provenance}
\end{table}

\paragraph{What the fix did not touch.}
Two facts bound the design exposure this leaves. The fix changed no
grading surface: no required-substring list, no claimed bound and no
tolerance (it did delete one agent-visible operating condition). And the
detector rules are structural predicates over tool presence, which are
hard to tune against leaked answers even in principle. What the fix does
not bound is the grading surface itself, since the confidence floor of
0.6 on three ACS70331 claims and the required-substring needles were
fixed inside the leak window and never revisited. We cannot rule out
that leaked runs shaped them, which is why
Appendix~\ref{sec:appendix-rederivation} re-derives the needles blind
(its floor arm was lost to ambiguous instructions).

\section{Claim List}
\label{sec:appendix-claims}

Each of the 25 claims records a parameter, an expected unit, a claim
kind, the stated operating conditions, and the chamber-side bindings
used to design its protocol.

\paragraph{Selection protocol.}
The candidate pool was the parameter tables of the three datasheets.
From those we kept quantitative parameters carrying an explicit value,
unit and operating condition; dropped any whose value cannot be stated
without reproducing a curve or a figure; and balanced the remainder
across claim kinds so that both verdicts are exercised rather than only
the cheaper one. The result is 14 typical performance characteristics,
6 operating ranges, 3 DC-accuracy specifications and 2 maximum ratings,
of which 5 have operating conditions the chamber can at least partly
reproduce (its \emph{chamber-realisable subset}). Every claim was verified
by hand against the datasheet's text layer, recording a source page and
a verbatim span. Two hand-authored artefacts then define the grading
surface: a required-substring list (19 claims carry two needles, 5 carry
three, 1 carries one) and a per-claim
confidence floor, 0.7 throughout
except for three ACS70331 claims at 0.6. Appendix~\ref{sec:appendix-rederivation}
re-derives the needles blind; its attempt at the floors failed.

Deployment relevance was a criterion but an informal one: we preferred
parameters an applications engineer is actually asked for, and we did
not record the rejected candidates. A third party can therefore re-apply
these criteria, but cannot reproduce our pool, and the same objection
applies with more force to the 12 claims of
Appendix~\ref{sec:appendix-fourth}, which have no declared pool at all.

\paragraph{Where the claims sit in each document.}
The claim file records a source page, not a section. The 25
claims cite 10 distinct pages; 12 name page~1, where a datasheet
puts its features and headline specifications, and only 2 rest on page~1
alone; the rest reach the parameter tables---pages 5 and 7--9 on the
DPS310, 4 and 47--52 on the Si115x, 5 on the ACS70331. The spread is a
property of how these documents are laid out, not a sampling design.

\paragraph{The gold set was revised after seeing model output.}
Four claims changed in response to runs, all of them before the reported
matrix. Three ACS70331 golds were \emph{corrections}, and the direction
matters: in each the curator was wrong and the datasheet settled it.
\texttt{acs70331-supply-current-max} was written as 7.5\,mA, which is a
different symbol's start-up current; the row reads 6\,mA.
\texttt{acs70331-saturation-low} conflated a typical column with a
maximum, and now records both. \texttt{acs70331-rise-time} was written
as 290\,ns against the 460\,ns the datasheet gives at the matching test
condition. All three carry the same hand-verification as every other
claim in the set---a source page and a verbatim span---so what the model
supplied was the discrepancy, not the value. The fourth claim was
\emph{dropped}: the Si115x datasheet states no linearity figure, and the
curator's threshold was extrapolated from the responsivity slope rather
than quoted, so there was no spec to score against.

This does not make fidelity a measure against the models' own answers:
the reported runs postdate every correction, and each is anchored in
the document. The limit it does leave is narrower: the audit was
\emph{model-triggered}, so a curator error the models happened to
reproduce would never have surfaced. Three of the five ACS70331 golds
needed correcting, which bounds how much confidence the unflagged ones
have earned.

\paragraph{A needle repaired after the fact.}
We found it only because the re-derivation re-scores the archived cells
under current code as a control. The \texttt{acs70331-saturation-low}
claim was gated on the symbol \texttt{VSAT\_LOW}, which appears nowhere
but the model's own quoted span; when that span was removed from the
matched text---so that quoting the right table row could no longer by
itself satisfy a value check---the needle became unsatisfiable, and the
claim's published verdict stopped reproducing under the code that had
produced it. We replaced it with the value and its unit, which pins the
same row (the adjacent \texttt{VSAT\_HIGH} is in volts, not
millivolts) and leaves the needle-count distribution and the numeric
needles unchanged. All 74
agentic cells that store an extraction reproduce their published verdict
under the repaired surface. One single-pass baseline cell does not
reproduce: the Qwen3.6-27B cell for this claim, the only one whose
quoted span omitted the symbol,
which therefore failed under the old needle and passes under the
repaired one. No figure in this paper moves.
The second annotator, blind to all of this, derived exactly the
replacement needle.

The chamber's bill of materials includes one further 20-page
motor-driver datasheet. The chamber cannot measure it, so it falls
outside the gradable set of Section~\ref{sec:design-claims}, but that
was never why it was left out: it was excluded on the grounds that it
adds no agentic-navigation stress beyond the three components above. We
withdraw that as a reason not to run it, and report the 12 claims we
did run on it in Appendix~\ref{sec:appendix-fourth}, where the
navigation measurements confirm the premise but not the conclusion.

\paragraph{What we publish.}
We release the benchmark in two tiers \citep{datasheetindex}. The first
is fully offline: the claim file, the grading surface, the archived runs
behind every number here, and the analyses that turn one into the other,
with no model client and no credentials. The second is the agent harness
that produced those runs---all three arms, a reference gateway
configuration, and a manifest naming the script and command behind each
archived artefact. The inputs were already public: the Causal Chambers'
open hardware, data and simulators, and manufacturer datasheets
identified by part number, which we cite rather than redistribute.
Both detector rules are predicates over the document-side tool layer's
API rather than over anything of ours, so they can be reimplemented from
that library alone, in a few dozen lines, without the harness.
What we cannot release is a model. Every arm is driven through a gateway
the replicator supplies, so a re-run reproduces our method exactly and
our numbers only up to the drift of the served checkpoints. Re-deriving
the grading surface (Appendix~\ref{sec:appendix-rederivation}) remains
the part we are least certain of.

\section{Why 23 Claims Are Not Physically Gradable}
\label{sec:appendix-inconclusive}

Section~\ref{sec:res-repro} counts the four classes. This appendix works
one claim from each, because each reason is a statement about the
apparatus: what a test laboratory would have to change.

\paragraph{Engagement-only (5 claims).}
All five are ACS70331 current-sensor claims, and the chamber stages that
component not at all: \texttt{acs70331-saturation-low} needs a supply
rail and a load resistance the apparatus never presents. A protocol can
be written and the agent can engage the claim, but no measurement exists
to compare against. The verdict is a statement about the bill of
materials rather than about the datasheet.

\paragraph{Absolute-accuracy, no traceable reference (5).}
\texttt{dps310-absolute-accuracy} asks whether the barometer is within
its stated absolute error. The chamber holds only its own sensors, so it
can show two of them agreeing and still certify neither; the unmatched
condition is named \texttt{external\_pressure\_reference}. Agreement is
not accuracy.

\paragraph{Unmatched decisive condition (12, the largest class).}
This class splits in two. Four claims need a stimulus the chamber cannot
apply: \texttt{si115x-supply-voltage-vdd} is specified over a VDD sweep,
and the apparatus powers its sensors from a fixed rail. The other eight
need a quantity it cannot observe: \texttt{si115x-standby-current-1v8}
is a device-internal supply current at 1.8\,V with no ADC or bus
activity, and the chamber measures neither the rail current nor the
quiescence. A claim in this class is not out of range; it is
unaddressed, which is why pooling it with a fail would be wrong.

\paragraph{Resolution-limited (1).}
\texttt{dps310-relative-accuracy} is staged, its conditions are matched,
and it still cannot be graded: the cross-sensor uncertainty is
$\approx$0.080\,hPa against a $\pm$0.06\,hPa tolerance. The apparatus is
the wrong instrument by a factor under two. This is the only class a
modest hardware upgrade would move.

\paragraph{Staging is not grading.}
Five claims carry a chamber-realisable subset---the ones a protocol can
actually stage---and four of those five still come back inconclusive:
the two worked in this appendix, \texttt{dps310-absolute-accuracy} and
\texttt{dps310-relative-accuracy}, plus
\texttt{dps310-pressure-precision-high} and
\texttt{dps310-pressure-data-resolution}. Only
\texttt{dps310-operating-range} reaches a definitive verdict. The second
definitive verdict, \texttt{si115x-adc-bit-depth}, is not in that set at
all: it has no chamber-realisable condition subset, its pass is an
output-code consistency check (the observed ADC codes fit the stated bit
depth) rather than an independent measurement of bit depth, and
Appendix~\ref{sec:appendix-rescore} reports that its fidelity pass rests
on model prose. So the envelope is narrower than the count suggests---of
the claims the apparatus can stage, it grades one.

\section{A Fourth Component, Off Corpus}
\label{sec:appendix-fourth}

Three components are a narrow base, and the natural objection is
breadth. The apparatus does not settle it: the constraint of
Section~\ref{sec:design-claims} binds only the claims a chamber must
\emph{grade}, and the chamber's bill of materials holds a fourth
datasheet it cannot measure, excluded for an unrelated reason we
withdraw in Appendix~\ref{sec:appendix-claims}. So we ran it: 12
claims were written on the 20-page A4988 stepper-driver datasheet, kept
apart from the frozen 25 and scored on fidelity only. It answers one
question---does the fidelity result, and the detector's silence, survive
a document outside the frozen corpus---and it answers it weakly.

\begin{table}[ht]
\centering
\small
\begin{tabular}{@{}lrrrr@{}}
\toprule
\textbf{Model} & \textbf{Fid.} & \textbf{Exact} & \textbf{Flags} & \textbf{Nav/claim} \\
\midrule
Claude Sonnet~4.6 & 12/12 & 12/12 & 0 & 5.25 (8.43) \\
GPT-5.1           & 12/12 & 12/12 & 0 & 7.25 (9.11) \\
\bottomrule
\end{tabular}
\caption{The fourth component, first run, no engine errors.
Navigation calls per claim are given with the model's own clean-corpus
mean in brackets. The comparison must be like for like: an all-tool
count against a navigation-only mean reverses the conclusion.}
\label{tab:fourth}
\end{table}

It is a weaker test than the frozen three, and we report it as a null
result rather than as generalisation. Five things bound it. The document
is navigated measurably more cheaply, $38\%$ and $20\%$ below each
model's corpus mean, which is what the original exclusion rationale
(Appendix~\ref{sec:appendix-claims}) predicted. The claims have no
declared candidate pool, so the selection criticism that applies to the
frozen 25 applies here with full force. Seven of their 18 numeric
needles are two characters or shorter---a $39\%$ share against $27\%$
for the frozen set---which makes a substring match easier to obtain by
accident. Only
one repeat per model was run here, against three on the frozen corpus,
so the within-configuration variance those repeats measure is not
measured for this component; the re-run below is a second configuration,
not a second repeat. And the PDF
is the same decoy the probe arm of Section~\ref{sec:res-silent} served,
so both models had been observed transcribing some of these rows before
the claims were written.

Zero detector flags over 12 claims per model is a zero-event count and
needs a stated unit of independence: a one-sided $95\%$ ceiling near
$22\%$ over the 12 claims, or near $12\%$ if the 24 cells were
independent, which they are not. It is not a precision result, because
with no positives in those cells precision has no denominator. The
oracle does not extend either: the chamber cannot stage a motor driver,
so all 12 are ungradable and the combined envelope is 2 of 37.

\paragraph{A re-run on a changed tool surface.}
The run above predates the harness's extracted-payload persistence, so
its cells can be re-run but not re-scored offline. We re-ran all 12
claims on both models four weeks later, and those cells do store the
payloads. Fidelity is unchanged: Claude passes 12 of 12 and GPT-5.1
11 of 11, under both matchers, with no detector flag on either model;
one GPT-5.1 cell (\texttt{a4988-logic-supply-current-max}) was lost to a
360\,s timeout.

Navigation counts across the two runs are \emph{not} comparable,
because the tool layer changed between them: \texttt{build\_datasheet}
began reporting where a table of contents came from, and because this
document's is reconstructed rather than read from bookmarks, the agent
is now told that its page numbers are inferred and advised to confirm a
section with full-text search before reading a range from it. The
\texttt{datasheetindex} release in use changed what that search
returns. Navigation duly rose
on GPT-5.1, on 10 of its 11 comparable claims and on none fewer. On
Claude the same surface change moved almost nothing---6 of 12 claims up,
3 down, 3 unchanged---so the effect is real on one model and absent on
the other, and neither run licenses a statement about drift.

The lesson for anyone instrumenting dispatch is that a tool-call count
is a property of the \emph{pair} (model, tool surface) and not of the
model, so the surface has to be versioned alongside the numbers;
otherwise an improvement to the tools reads as a regression in the
model. The corpus means in Table~\ref{tab:fourth} were measured on the
earlier surface, which is why the like-for-like comparison is the one
reported there. What the re-run adds is that the fidelity result and
the absence of detector flags survive a change to the tool surface
underneath them.

\section{Blind Re-Derivation of the Grading Surface}
\label{sec:appendix-rederivation}

Every fidelity number in this paper is measured with a grading surface
that two hand-authored artefacts define: the required-substring list
attached to each claim, and the per-claim confidence floor. Both were
fixed inside the leak window and never revisited
(Appendix~\ref{sec:appendix-audit}). Re-running the models does not
address this, because the surface they are graded against is the thing
in question.

A second annotator therefore re-derived both from the datasheets alone.
They worked from the claim text, the component datasheet and the
protocol description, were blind to the current needle lists and floors
and to every run, and had no part in authoring either artefact. We
report the agreement per claim and in aggregate, each disagreement, and
whether it would change a published verdict.

\paragraph{Needles.}
The annotator answered 23 of the 25 claims and abstained on two.
Comparing the needle lists as written, the numeric needles are identical
on 15 of 23 and the unit needles on 13; no claim matches on needles and
floor together. Most of that gap is notation rather than disagreement.
The annotator wrote range ends as \texttt{'min 1.7'} where we write
\texttt{'1.7'}, kept the $\pm$ and $<$ glyphs we drop, and wrote units
more strictly than we do---\texttt{'Pa RMS'} for our \texttt{'Pa'},
\texttt{'$\mu$s'} for our \texttt{'s'}. Normalising those conventions,
the numbers agree on 21 of those 23 answered claims. Of the two that
remain,
\texttt{dps310-absolute-accuracy} is $\pm100$\,Pa against our
$\pm1$\,hPa, the same quantity at a different scale. One genuine
disagreement survives in 23 claims. In sum, over the frozen set the two
derivations agree on 21 of 25 claims, differ on 2 and leave 2
unanswered; we take the stricter reading and count the unit-scale
restatement as a difference.

The needles are not a separate artefact from the claimed values: on 24
of the 25 claims a numeric needle is one of the claim's own declared
bounds, so agreeing on the needles is agreeing on the number. The
exception is \texttt{si115x-adc-bit-depth}, whose needle is the bit
depth while its declared bounds are the converter's output range---the
same mismatch Appendix~\ref{sec:appendix-rescore} reports for that
claim.

The unit gap runs against us: on those claims the annotator's needles
are the stricter ones. Ours include
single-character units such as \texttt{'s'} and \texttt{'m'}, the same
weakness Appendix~\ref{sec:appendix-rescore} reports for two-character
unit needles matching inside compound tokens.

\paragraph{A residual leak in the guide.}
The guide shipped with the blind bundle used \texttt{'1.7'} as its
formatting example---the first needle of
\texttt{dps310-supply-voltage-vdd}---so one claim of 25 was not strictly
blind, on its numeric needle only. The record bounds it: the example read
``write \texttt{'1.7'}, not \texttt{'min 1.7'}'', and the annotator wrote
\texttt{'min 1.7'}. An earlier version of this guide leaked four answers
and was fixed before the round began; this one survived that fix, and we
found it only while preparing the release.

\paragraph{The disagreement, which is an annotator error.}
On \texttt{si115x-vis-response-525nm} the annotator recorded 125 where we
record 160. Table 8.3 of the Si115x datasheet gives
160\,counts/(W/m$^2$) for the small visible photodiode at 525\,nm under
\texttt{ADCMUX=0x11}; 125 does not appear in that table, and is the value
of a different claim on the same datasheet. We read it as a
transcription slip rather than a competing reading of the row. We
report it rather than correcting it: a second derivation that agreed
everywhere would be weaker evidence of independence, not stronger.

\paragraph{Abstentions.}
Both are informative and one of them is ours. On
\texttt{dps310-relative-accuracy} the annotator declined because the
claim records a 3.0\,V supply while the datasheet specifies the parameter
at 1.8\,V, on every page that carries the relevant tables. They are
right. The claim's condition list mixes chamber bindings with datasheet
conditions without marking which is which, so the abstention is a
defect in our form, not in their reading. The second, \texttt{dps310-operating-range}, is a miss:
the range is stated on the datasheet's first page.

\paragraph{Floors.}
The floors could not be independently re-derived, because our
instructions were ambiguous: the annotator used the field to express
confidence in their own derivation, where we intended a threshold on
the confidence the agent reports for itself. We report this arm as lost
rather than as disagreement. The annotator set 1.0 on 19 of the 23
answered claims, left the field blank on two, and departed downward on
two, one of them recording doubt about their own reading in a note.

The fitting question the floor raises can still be answered, from the
archive alone and more directly than agreement would have answered it.
The floors take two values, 0.7 on 22 claims and 0.6 on three, and are
not tuned per claim. Against them, across the 74 agentic cells that
store an extraction, the agent's self-reported confidences clear the
floor by a median of 0.27; only one of the 73 that pass sits within 0.05
of its floor, and the floor is decisive for exactly one cell.
A gate standing open by that margin cannot have been fitted to admit a
particular answer. The one cell it does decide is
\texttt{si115x-adc-bit-depth} on Qwen3.6-27B, one of that model's two
failures---and the same claim the annotator marked as the one they were
least sure of.

\paragraph{Effect on published verdicts.}
The re-derived floors are not a valid alternative surface, since the
annotator read the field differently from us (above); we apply them
mechanically here to show the consequence, not to propose a corrected
score. Applied literally to the archived extractions, the re-derived
surface moves 64 of 68 testable cell verdicts, every one from pass to
fail.
Attributing each flip to its first sufficient cause: 51 follow from the
1.0 floor, which rejects every answer the agent has ever given; 3 from a
stricter unit needle; 4 from the annotator's \texttt{min}/\texttt{max}
range prefixes, which the substring matcher enforces literally even
though the paragraph above treats them as notation; and 6 from the two
numeric differences, of which 3 are the unit-scale restatement and 3 the
single annotator error. That ordering matters, because the causes are
not disjoint: 24 of the 64 cells fail the floor and a needle at once,
and varying the unit needles alone, holding our own numerics and floors,
moves 13 cells rather than 3. No published figure rests on either
disputed \emph{numeric} value: the unit-scale restatement is the same
quantity, and the single annotator error is theirs, not ours.

\section{Latency Decomposition}
\label{sec:appendix-latency}

Section~\ref{sec:res-cost} says latency is dominated by time outside
local tool execution, and this appendix gives the split behind that. Over the post-audit baseline
run, local tool execution accounts for 10.4\,s of a 72.0\,s wall-clock
mean per cell on Claude Sonnet~4.6 ($14.4\%$), 12.2\,s of 146.9\,s on
GPT-5.1 ($8.3\%$), and 9.1\,s of 127.1\,s on Qwen3.6-27B ($7.2\%$).

The remainder is model generation plus gateway queueing, which these
traces cannot separate: a provider turn is recorded as one span. Two
caveats: this is a single run per model, so the values carry no
interval; and one cell had two attempts with identical tool counts, an
arbitrary tie-break worth at most 0.2\,s.

The agentic premium is therefore not the tool layer: even where local
execution is the largest share, seven wall-clock seconds in eight are
spent outside our control.

\section{Production Pilot Audit}
\label{sec:appendix-pilot}

Section~\ref{sec:intro} describes an internal pilot. This appendix
states what dispatch instrumentation actually covers there.

The pilot ran 126 extraction jobs between 2026-04-02 and 2026-07-22.
Dispatch instrumentation was deployed on 2026-06-11, part-way through,
and covers 17 of those jobs; one job produced two extraction records,
so 18 extractions. Of those, 14 jobs (15 extractions) routed as
large-PDF and are therefore eligible to raise either rule at all. Two
of the remaining three are small-PDF jobs, which register no navigation
tools and never flag in any engine. The third recorded no routing line,
so its eligibility is unknown and we count it in neither direction.

No rule has fired organically: 0 of 14 eligible jobs, 0 of 15
extractions. A one-sided $95\%$ ceiling on 0 of 14 is $19\%$, and even that
overstates what we know, because those jobs re-use documents and
configurations and are not independent trials. Both silent failures this
paper reports, the zero-tool-call run of Section~\ref{sec:worked-example}
and the dropped reasoning trace of
Appendix~\ref{sec:appendix-portability}, were found by a human reading
a trace, not by a rule firing.

One further measurement bears on the deployed source-grounding pass
of Section~\ref{sec:res-silent}. Across three audited runs,
17 of 84 extracted values ($20\%$) could not be re-located in their
source document, and no flag was raised on any of them: the check flags
only a locator error.

\section{Probe Arms and Their Coverage}
\label{sec:appendix-probes}

Three arms corrupt an agent's access to the document while leaving the
claim, the task and the scoring untouched. \emph{Closed-book} registers
no document tool at all, so an answer can only come from parametric
memory. \emph{Null-tool} registers the full surface and returns empty
content from every call. \emph{Wrong-content} serves every call from a
decoy datasheet, the A4988 motor driver, so navigation, cross-check and
read success are all genuine and only the document is wrong. The last
two are the tools-callable arms of Section~\ref{sec:res-silent}: both
satisfy the cross-check the detector requires, so both are invisible to
it by construction.

\begin{table}[t]
\centering
\small
\setlength{\tabcolsep}{4pt}
\begin{tabular}{@{}llrrrr@{}}
\toprule
\textbf{Arm} & \textbf{Model} & \textbf{Compl.} & \textbf{Err.} & \textbf{Ans.} & \textbf{Right} \\
\midrule
Closed-book   & Claude   & 25 & 0  & 17 & 7 \\
              & GPT-5.1  & 25 & 0  & 2  & 1 \\
              & Qwen     & 19 & 6  & 5  & 0 \\
\midrule
Null-tool     & Claude   & 25 & 0  & 1  & 0 \\
              & GPT-5.1  & 25 & 0  & 1  & 0 \\
              & Qwen     & 8  & 17 & 0  & 0 \\
\midrule
Wrong-content & Claude   & 25 & 0  & 2  & 0 \\
              & GPT-5.1  & 25 & 0  & 4  & 0 \\
              & Qwen     & 7  & 18 & 0  & 0 \\
\bottomrule
\end{tabular}
\caption{Probe arms, 25 claims per model and arm. ``Ans.'' counts runs that
returned a value rather than declining; ``Right'' counts those that also
passed fidelity. Qwen3.6-27B is scored for closed-book only; its two
tools-callable rows are shown for completeness but lose most runs to
engine errors.}
\label{tab:probes}
\end{table}

\paragraph{Why Qwen is excluded from two of the three.}
Our harness forces tool choice. Closed-book registers no tools, so the
forcing cannot loop and the arm is valid on all three models. The other
two give it a surface to loop on: Qwen3.6-27B errors on 17 and 18 of 25
runs, dominated by token exhaustion, and on the runs that do complete
its mean navigation count is 45 against 9.5 and 21.6 for the other two
models (Table~\ref{tab:probes} and
Appendix~\ref{sec:appendix-qwen-thinking}). Those arms would measure the
forcing rather than the model, so we report them on Claude Sonnet~4.6
and GPT-5.1 only, and the 100-run denominator in
Section~\ref{sec:res-silent} is two arms $\times$ two models $\times$ 25
claims. The exclusion is a coverage limit, not a result: nothing here
says Qwen would or would not answer from a decoy.

\paragraph{What closed-book establishes.}
With no document at all, Claude Sonnet~4.6 answers 17 of 25 claims and
is right on 7 of them ($41\%$); GPT-5.1 answers 2 and is right on 1;
Qwen answers 5 and is right on none. Recovery concentrates on the
DPS310, the most widely documented of the three components---6 of its 11
claims, against 1 of 9 on the Si115x and 0 of 5 on the ACS70331---which
is the shape of memorised public documentation rather than of reading.
Every one of Claude's 7 recoveries carries a fabricated quotation: the
channel on which, as Appendix~\ref{sec:appendix-rescore} shows, two
fidelity passes rest.

\section{Exact-Value Re-Score}
\label{sec:appendix-rescore}

Fidelity is scored by requiring each claim's needles to appear in the
extraction. Substring matching is what the headline uses, and it is
lenient: a needle of ``1'' matches inside ``1.7''. We therefore re-scored
every recoverable cell with numeric needles compared numerically rather
than by substring.

It did not move the headline. Agentic fidelity is 25/25 for Claude
Sonnet~4.6, 25/25 for GPT-5.1 and 23/25 for Qwen3.6-27B under both
matchers. Coverage is 73 of the 207 clean cells: only the reference run
(Appendix~\ref{sec:appendix-cells}) stores extractions, so repeats 2 and 3 are verdict-only and cannot be
re-scored offline at all.

Five things it exposed are worth more than the headline.

\paragraph{A permutation control.} Pairing each claim with a
\emph{different} cell's extraction, 13 of 600 mismatched pairs pass on
Claude ($2.2\%$) and 17 of 600 on GPT-5.1 ($2.8\%$). The low rate
shows the needles discriminate. But the leakage concentrates in
weak-needle claims, so the caveat belongs on figures resting mainly on
those: not the headline, but the closed-book recovery rate of
Appendix~\ref{sec:appendix-probes}.

\paragraph{Passes resting only on model prose.} Two of the 73
re-scorable cells satisfy a numeric needle nowhere in the structured
value fields, only inside the model's own quotation: the same claim
(\texttt{si115x-adc-bit-depth}) on both models, whose structured fields
hold the converter's output range rather than its bit depth. That
channel is the one the closed-book arm
(Appendix~\ref{sec:appendix-probes}) shows can be fabricated: every
recovery it produced carried a quotation found in no datasheet. It also
happens to be one of the two claims reaching a definitive chamber
verdict, so our second definitive physical result rests on a fidelity
pass grounded in prose alone.

\paragraph{The closed-book rate, re-scored.} The recovery figure of
Appendix~\ref{sec:appendix-probes} is a substring result, so we
re-score it too. Under numeric matching
Claude's recoveries fall from 7 of 17 answered ($41\%$) to 6 of 17
($35\%$), the single loss being a missing unit; across all three models
recoveries go from 8 to 7. The prevalence gradient is unchanged
(5 of 11 on the DPS310, 1 of 9, 0 of 5).

\paragraph{The two figures are not a pair.} A reader meeting $25/25$
here and 8 recoveries there should not divide one by the other. The
benchmark cells are scored over the whole extracted record; the
closed-book cells are scored over the submitted quotation alone, because
that is all an arm with no document produces. The matchers are the same;
the surfaces they run over are not.

\paragraph{Unit needles matching inside longer tokens.} Five passes
match a unit needle inside a compound token: \texttt{Pa} inside
\texttt{PaRMS}, \texttt{mV} inside \texttt{mVpp}. Both are correct
readings, but the matcher cannot tell them from a wrong unit that
happens to be a prefix, and we report the count rather than assert that
every such match is safe.

\section{Cell Populations}
\label{sec:appendix-cells}

A \emph{cell} is one (model, claim, repeat) extraction
(Section~\ref{sec:setup}), but the paper reports over several
populations of them, and two of the counts collide.
Table~\ref{tab:cells} lists them.

\begin{table}[ht]
\centering
\small
\setlength{\tabcolsep}{4pt}
\begin{tabular}{@{}rll@{}}
\toprule
\textbf{N} & \textbf{Population} & \textbf{Reported in} \\
\midrule
75  & agentic cells, one repeat        & \S\ref{sec:res-cost} \\
74  & \hspace{1.2em}storing an extraction        & App.~\ref{sec:appendix-claims} \\
73  & \hspace{1.2em}and fidelity-passing         & App.~\ref{sec:appendix-rescore} \\
\midrule
207 & fidelity-passing, three repeats  & \S\ref{sec:res-silent} \\
74  & \hspace{1.2em}claim-by-model groups        & \S\ref{sec:res-silent} \\
\bottomrule
\end{tabular}
\caption{The cell populations this paper reports over. Each block
nests within itself; the two 74s are unrelated counts that happen to
coincide.}
\label{tab:cells}
\end{table}

The single-repeat population is the post-audit \emph{reference run}
(repeat~1 of the agentic suite; not the single-pass baseline engine):
25 claims $\times$ 3 models $=$ 75 agentic cells, the denominator for the
baseline-versus-agentic comparison of Section~\ref{sec:res-cost}. Of
those, 74 store the model's extraction rather than its verdict
alone---the harness gained payload persistence part-way
through---and 73 of the 74 pass fidelity. The 73 is what
Appendix~\ref{sec:appendix-rescore} can re-score offline; the 74 is what
Appendices~\ref{sec:appendix-claims}
and~\ref{sec:appendix-rederivation} re-check against a repaired or
re-derived grading surface.

The three-repeat population is the variance suite: 207 fidelity-passing
agentic cells, 75 on Claude Sonnet~4.6, 75 on GPT-5.1 and 57 on
Qwen3.6-27B. It is the denominator for the detector's false-positive
result. Its repeat~1 \emph{is} the reference run, so the two populations
must never be added (their sum was a figure this paper retracted).

The collision to watch is that 74 names two different things. One is
the 74 baseline cells that store an extraction. The other is the 74
claim-by-model groups inside the 207 ($25+25+24$, one short of
$25\times3$ because \texttt{si115x-adc-bit-depth} never passed on
Qwen3.6-27B in any repeat). Neither count derives from the other.

\section{Long-Datasheet Stress Test}
\label{sec:appendix-stress}

The benchmark corpus tops out at 65 pages, where the tool layer is pure
overhead (Section~\ref{sec:res-cost}). To exercise the opposite regime we
ran both engines on a 397-page automotive PMIC datasheet (TI TPS6594-Q1)
across all three models. How far the single-pass baseline reaches depends
on its ingestion path: the native PDF document block is rejected
outright (Claude's gateway caps a PDF at 100 pages, and GPT-5.1's
request overflows the context window), while Qwen's rendered-page-image
baseline does ingest all 397 pages, but only by spending 208k input
tokens in one call, a brute-force cost that scales with every page. The
agentic engine sidesteps both walls: reading only the pages each claim
needs, each of the three models extracts all five target parameters at
0.93 confidence or above, in 83--139\,s per model, every value exact
against the datasheet's electrical-characteristics tables. The tool layer is overhead
when the document fits and bounded-cost where single-pass ingestion is
not; we compare it against that one alternative, not against chunked
retrieval or a map-reduce pass, which we did not run. A corpus that only
ever exercises the first regime would mistake that overhead for waste.

One caveat. The stress harness reports to standard output and archives
nothing, so the figures above come from a single exploratory run with no
stored per-cell record: they cannot be re-scored offline as the
benchmark's own cells can, and the page count is the only one of them a
reader can check independently. The document-side tool surface has moved
since as well (Appendix~\ref{sec:appendix-fourth}), so a fresh run would
not be a like-for-like repeat.

\section{Qwen Reasoning-Mode Tool-Call Interaction}
\label{sec:appendix-qwen-thinking}

The 23\% per-repeat engine-error rate on Qwen3.6-27B in
Table~\ref{tab:results} arises from a documented interaction between
vLLM's chat-template handling of reasoning mode
(\texttt{enable\_thinking=true}) and structured tool-call emission
\citep{qwen3issue1817}: the model plans its submit call
(\texttt{submit\_claim\_result}, Section~\ref{sec:worked-example}'s
predecessor tool) within its thinking block, satisfies itself by
reasoning about the action, then ends the turn
(\texttt{stop\_reason=end\_turn}) without emitting the tool-call tokens.
Our agentic loop sees a terminal turn carrying no \texttt{tool\_use}
block and raises \texttt{SubmitToolNotCalledError}, the same path the
harness already uses for max-turn exhaustion. Because these repeats
predate the two-pass freeze (Section~\ref{sec:setup}), every Qwen cell
that submits at all does so through that predecessor tool, where the
frontier models' cells call \texttt{submit\_extraction}; the tool
surface differs with the freeze, not with the model.

To isolate this configuration-level effect from intrinsic model
behaviour we re-ran the variance experiment with
\texttt{enable\_thinking=false}, keeping the prompts, tools, gateway,
and 360\,s per-cell timeout fixed.
Table~\ref{tab:qwen-thinking} compares the two configurations across
three fresh repeats of the same 25-claim suite.

\begin{table}[ht]
\centering
\small
\setlength{\tabcolsep}{4pt}
\begin{tabular}{@{}lcc@{}}
\toprule
\textbf{Per-repeat metric} & \textbf{Reas.\ ON} & \textbf{Reas.\ OFF} \\
\midrule
Fidelity / 25            & 23 / 19 / 15 & 25 / 24 / 24 \\
Engine errors            & 1 / 6 / 10   & 0 / 0 / 1    \\
Engine-error rate        & 23\%         & 1.3\%        \\
Mean latency (s)         & 130          & 64           \\
Stable claims / 25       & 13           & 24           \\
\bottomrule
\end{tabular}
\caption{Qwen3.6-27B variance under the two reasoning-mode
configurations, three repeated runs of the same 25 claims. Disabling
reasoning cuts the engine-error rate by an order of magnitude and
roughly halves per-cell latency while preserving fidelity. The single
residual engine error (\texttt{si115x-adc-bit-depth}, max-turn
exhaustion) is a different failure mode (tool over-triggering rather
than tool-call drop) and matches the secondary effect that
\citet{qwen3issue1817} reports for the \texttt{enable\_thinking=false}
workaround.}
\label{tab:qwen-thinking}
\end{table}

Both configurations are realistic deployments: reasoning-on is the
default a practitioner reaches for to exploit Qwen3's published
reasoning mode; reasoning-off is the workaround currently recommended
in the upstream issue. We retain the reasoning-on configuration for
the Section~\ref{sec:results} headline numbers; the value of this
ablation is not which number is ``correct'' but that the benchmark's
per-cell instrumentation surfaces a configuration-level failure mode
that aggregate fidelity scoring would hide.

\section{Positioning and Additional Figures}
\label{sec:appendix-figures}

\begin{table}[t]
\centering
\small
\resizebox{\columnwidth}{!}{%
\begin{tabular}{@{}lll@{}}
\toprule
\textbf{Work} & \textbf{Oracle} & \textbf{Domain} \\
\midrule
ModelGen        & MLLM judge      & Semiconductor models \\
AMSbench        & Mixed           & AMS circuits \\
DocDancer       & Doc-internal    & Long-context docs \\
DCI             & QA correctness  & Web / multi-hop QA \\
RWE-bench       & Medical DB      & Clinical evidence \\
Causal Chambers & Physical meas.\ & Causal discovery \\
\midrule
This paper      & Physical meas.\ & Component datasheets \\
\bottomrule
\end{tabular}%
}
\caption{Positioning against related benchmarks. Ours is, to our
knowledge, the first to pair document-extraction fidelity with
independent physical measurement of the same claims
(Section~\ref{sec:res-repro}).}
\label{tab:related}
\end{table}

\begin{figure}[ht]
\centering
\includegraphics[width=\columnwidth]{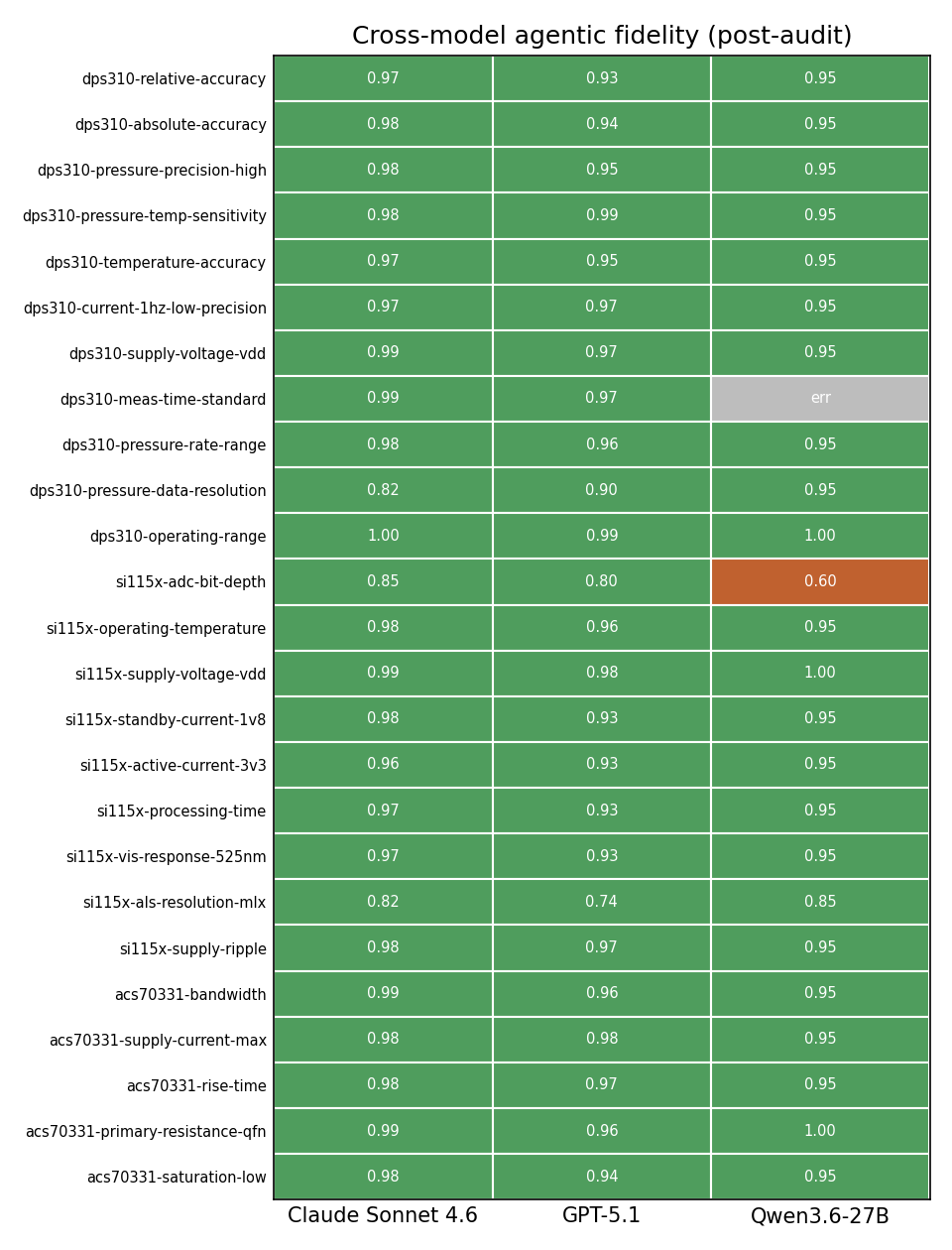}
\caption{Per-claim agentic fidelity by model (post-audit). Green is a
fidelity pass, orange a fail, grey an engine error; the number is the
per-claim confidence.}
\label{fig:fidelity}
\end{figure}

\begin{figure}[ht]
\centering
\includegraphics[width=\columnwidth]{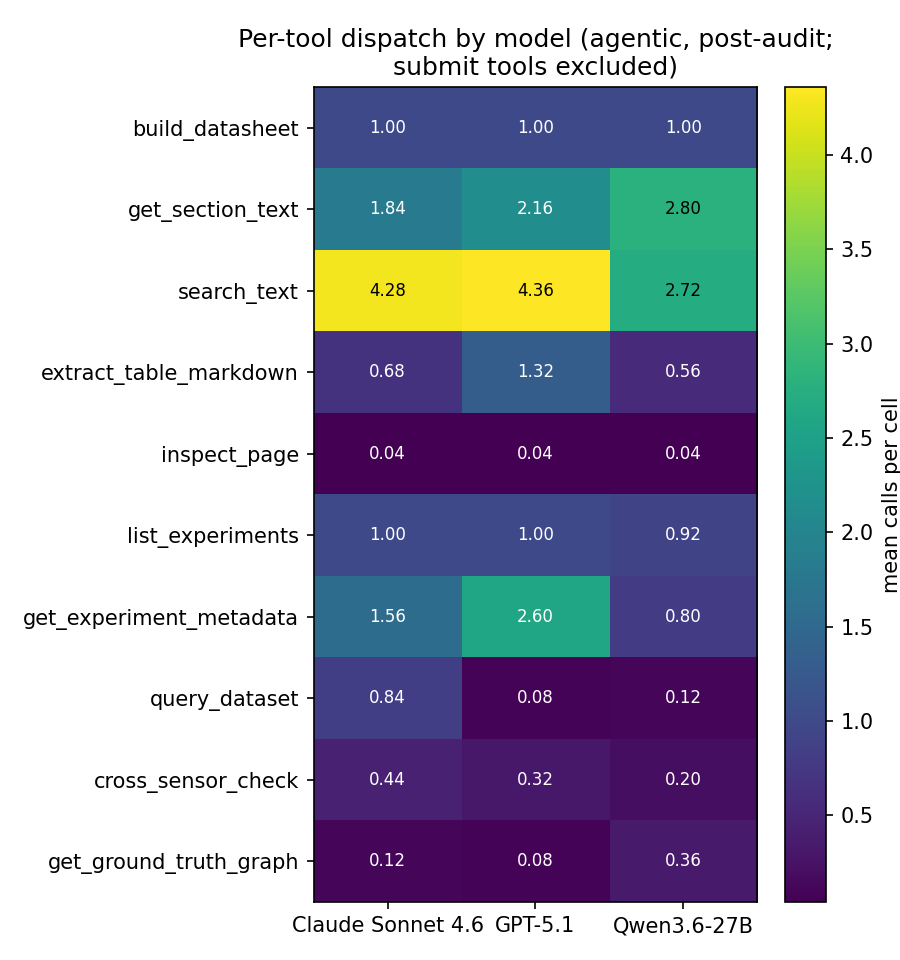}
\caption{Mean calls per cell for each tool, by model
(agentic engine, post-audit). The means cover the document-side tools
only: submit tools and the entire chamber phase are excluded, so these
are not the counts the cross-check predicate of
Section~\ref{sec:res-silent} reads against a chamber-side call.
All three models exercise the tool surface; this dispatch record is
what made the silent zero-tool-call failure of
Section~\ref{sec:worked-example} detectable.}
\label{fig:dispatch}
\end{figure}

\begin{figure}[ht]
\centering
\includegraphics[width=\columnwidth]{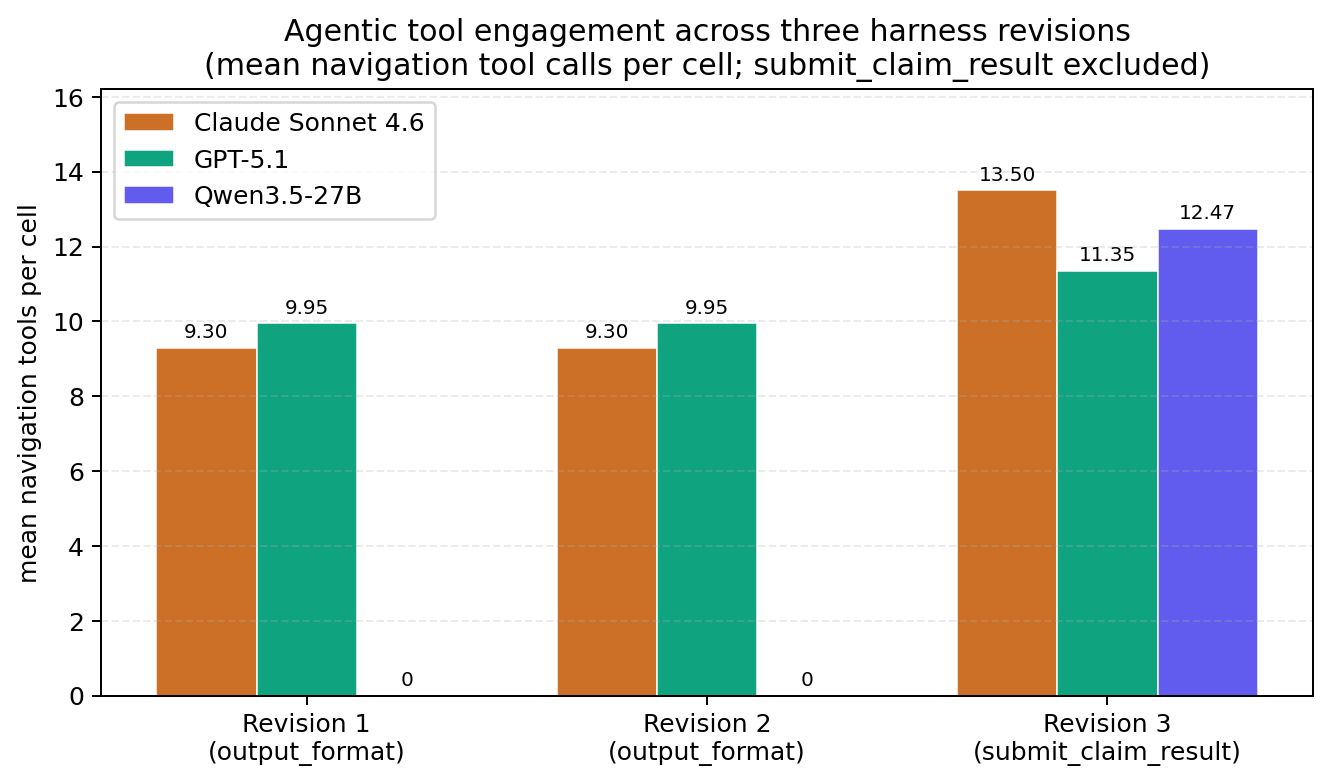}
\caption{Mean navigation-tool calls per cell across three development
revisions. Qwen issues zero tool calls until the submit-tool channel
replaces the \texttt{output\_format} constraint, after which it engages
the datasheet---the silent tool-bypass of
Section~\ref{sec:worked-example} and its fix. This development period
predates the Qwen3.5-to-3.6 gateway swap, so the figure shows
Qwen3.5-27B; the cross-model results in Section~\ref{sec:results} use
Qwen3.6-27B. It also predates the oracle-leak fix
(Appendix~\ref{sec:appendix-audit}): the first two revisions are
pre-audit measurements carried forward as constants, and only the third
is recomputed from the post-audit matrix. Nothing here is a fidelity
number---the axis is navigation calls---but the boundary is worth
naming, since Section~\ref{sec:design-agent} says every result in
Section~\ref{sec:results} is post-audit.}
\label{fig:engagement}
\end{figure}

\begin{figure}[ht]
\centering
\includegraphics[width=\columnwidth]{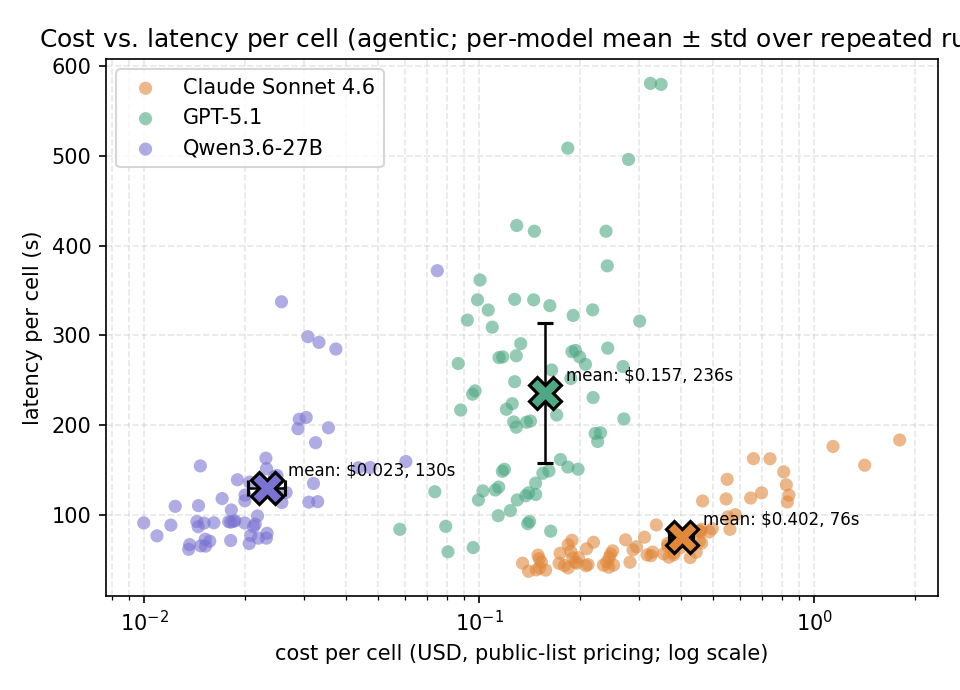}
\caption{Cost and latency per cell under the agentic engine, by model
(cost on a log scale; large markers are per-model means, with $\pm$ std
error bars over the three repeated runs). At list prices agentic
extraction costs about \$0.38 per claim for Claude and \$0.16 for
GPT-5.1, $1.2\times$ and $1.8\times$ their single-pass baselines;
Qwen's $3.9\times$ ratio comes from a cheap baseline rather than an
expensive agent, its page-image input being $\approx$24k tokens against
Claude's $\approx$98k. Fidelity is close across these points: the models
separate far more sharply on cost and latency.}
\label{fig:cost}
\end{figure}

\begin{figure}[ht]
\centering
\includegraphics[width=\columnwidth]{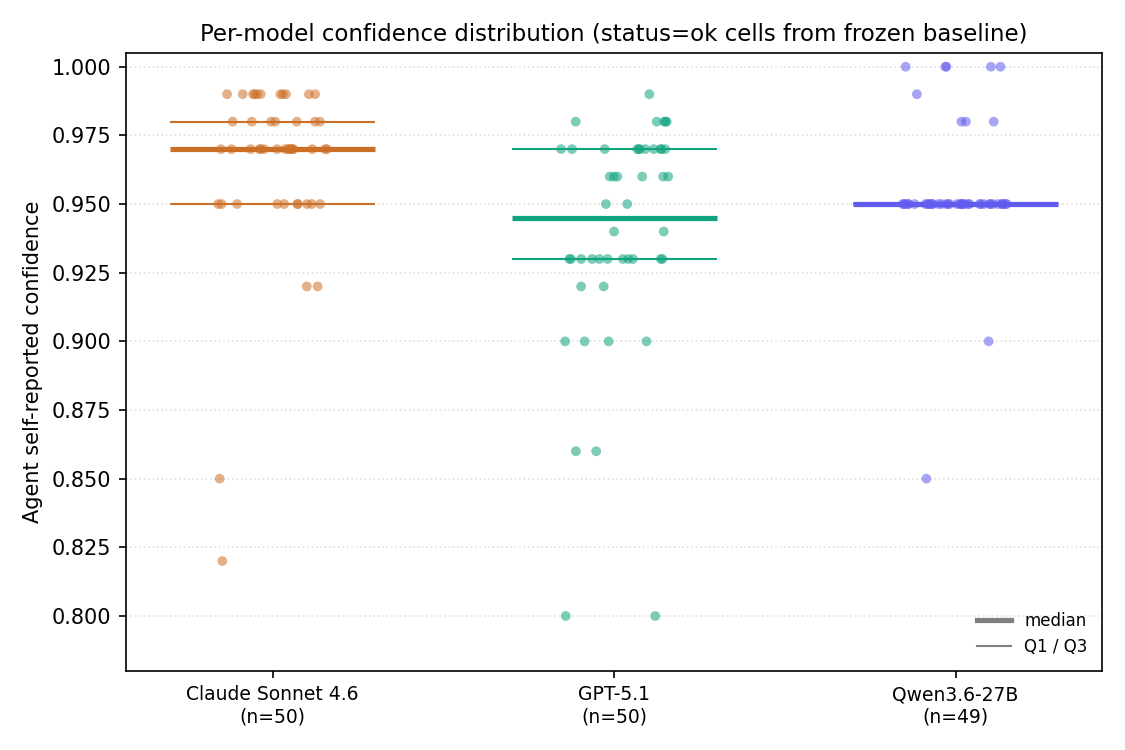}
\caption{Per-model distribution of agent-reported confidence over the
status-ok cells of the post-audit baseline; medians and quartiles
marked.}
\label{fig:confdist}
\end{figure}

\end{document}